\documentclass[10pt,twocolumn,letterpaper]{article}

\usepackage{cvpr}
\usepackage{times}
\usepackage{epsfig}
\usepackage{graphicx}
\usepackage{amsmath}
\usepackage{amssymb}
\usepackage{comment}
\usepackage{subcaption}
\usepackage{algorithm}
\usepackage{algorithmic}

\usepackage[pagebackref=true,breaklinks=true,letterpaper=true,colorlinks,bookmarks=false]{hyperref}

\cvprfinalcopy 

\def\cvprPaperID{2871} 

\ifcvprfinal\fi
\begin{document}

\title{Photometric Stereo in Participating Media Considering Shape-Dependent Forward Scatter}

\author{Yuki Fujimura$^1$
\and
Masaaki Iiyama$^1$
\and 
Atsushi Hashimoto$^{1,2}$
\and
Michihiko Minoh$^1$
\and
$^1$Kyoto University, Japan
\and
$^2$OMRON SINIC X Corp., Japan
\and
{\tt\small \{fujimura, iiyama, a\_hasimoto, minoh\}@mm.media.kyoto-u.ac.jp}
}

\maketitle
\thispagestyle{empty}

\begin{abstract}
Images captured in participating media such as murky water, fog, or smoke are degraded by scattered light. Thus, the use of traditional three-dimensional (3D) reconstruction techniques in such environments is difficult. In this paper, we propose a photometric stereo method for participating media. The proposed method differs from previous studies with respect to modeling shape-dependent forward scatter. In the proposed model, forward scatter is described as an analytical form using lookup tables and is represented by spatially-variant kernels. We also propose an approximation of a large-scale dense matrix as a sparse matrix, which enables the removal of forward scatter. Experiments with real and synthesized data demonstrate that the proposed method improves 3D reconstruction in participating media.
\end{abstract}

\section{Introduction}
\begin{figure}[tb]
  \centering
  \begin{minipage}[b]{0.49\hsize}
    \centering
    \includegraphics[scale=0.25]{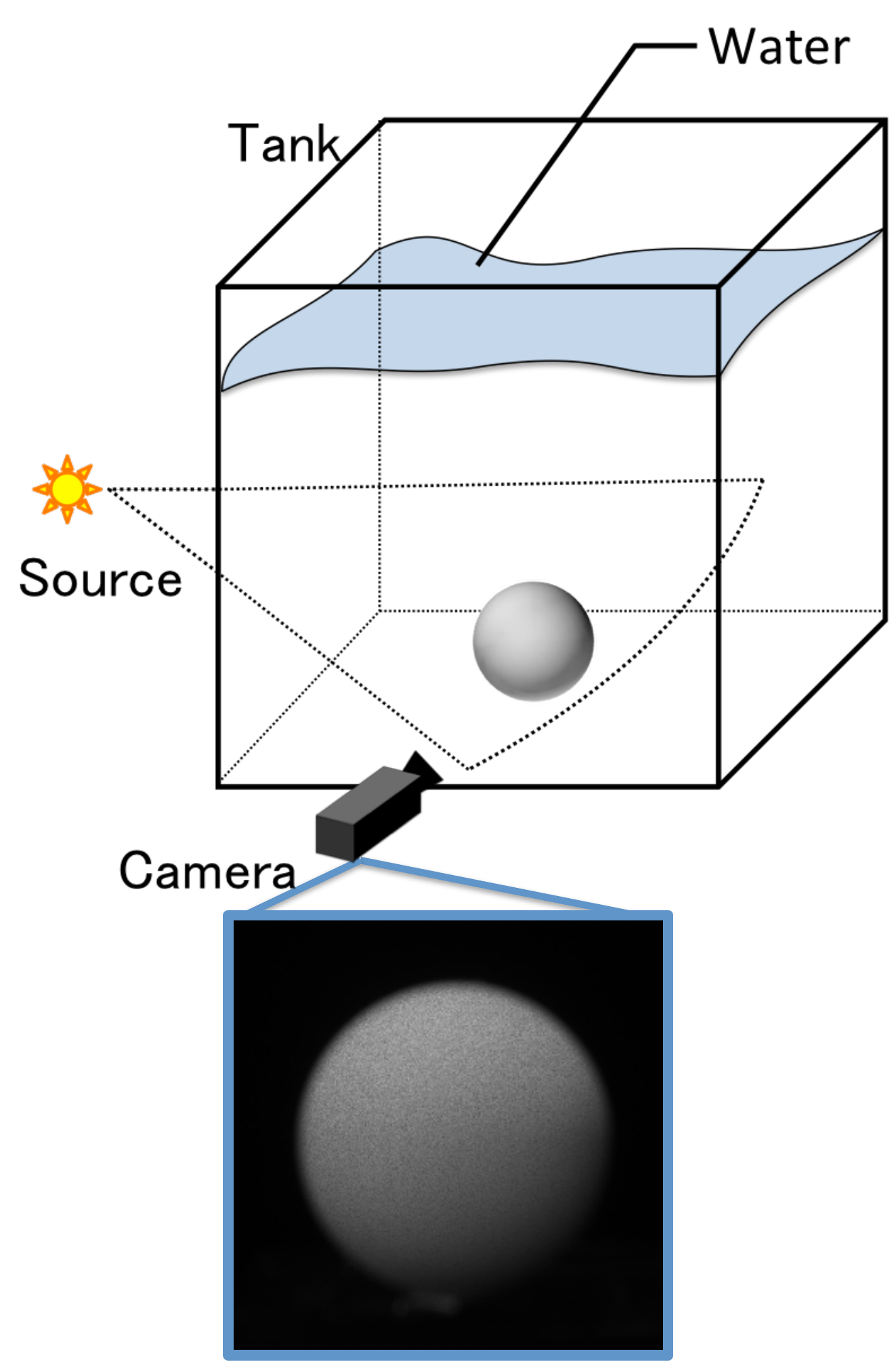}
    \subcaption{}
    \label{fig:sphere_in_water}
  \end{minipage}
  \begin{minipage}[b]{0.49\hsize}
    \centering
    \includegraphics[scale=0.25]{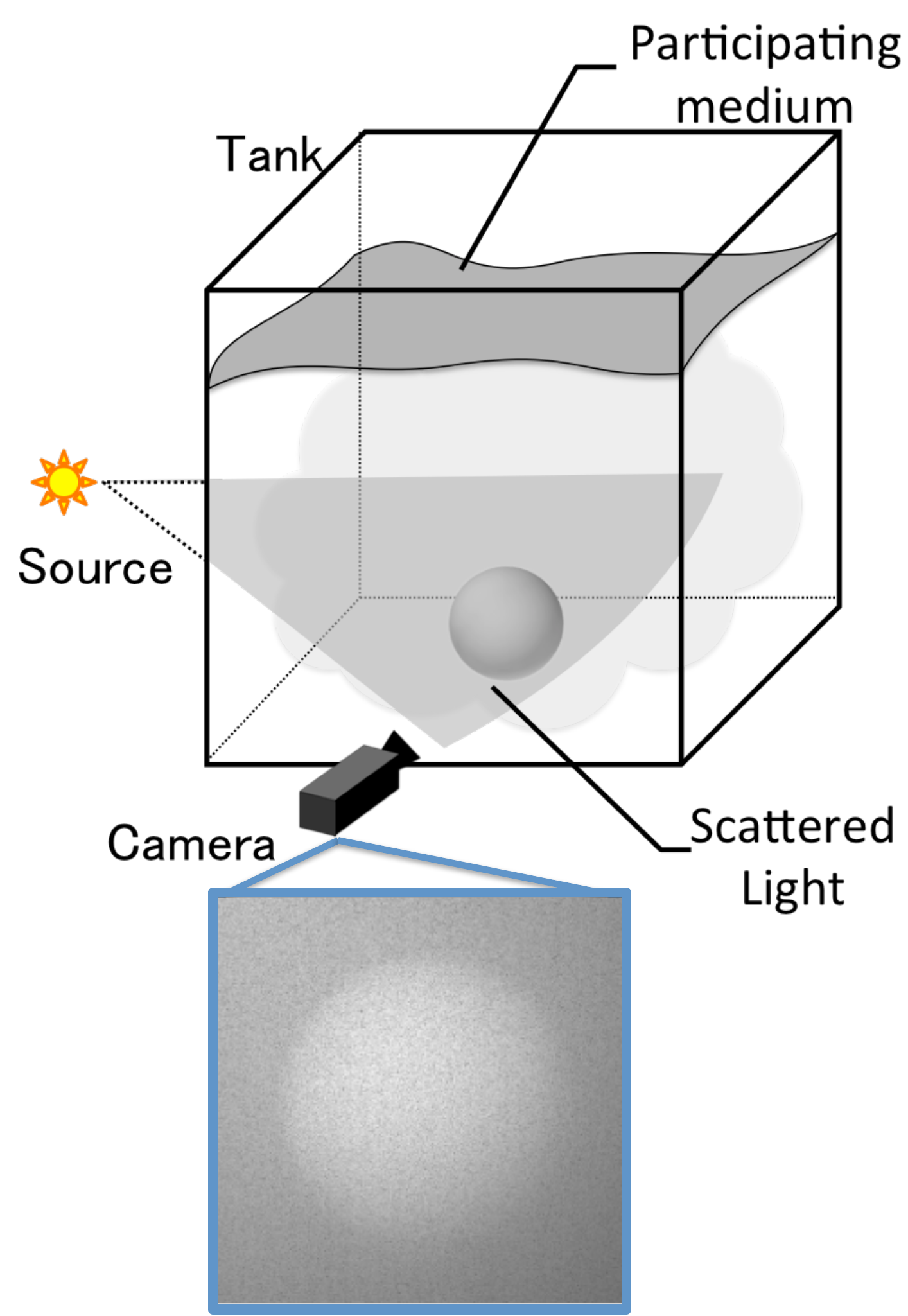}
    \subcaption{}
    \label{fig:sphere_in_media}
  \end{minipage}
  \caption{Images captured in (a) pure water and (b) diluted milk. In participating media, the quality of the captured image is degraded by light scattering and attenuation. }
  \label{fig:degraded}
\end{figure}

Three-dimensional (3D) shape reconstruction from two-dimensional (2D) images is an important task in computer vision.
Numerous 3D reconstruction such as structure from motion, shape-from-X, and multi-view stereo have been proposed. 
However, reconstructing the shape of an object in a participating medium, \eg, murky water, fog, and smoke, remains a challenging task.
In participating media, light is attenuated and scattered by suspended particles, which degrades the quality of the captured images (Figure \ref{fig:degraded}). 
3D reconstruction techniques designed for clear air environments will not work in participating media.

\begin{figure}[tb]
  \centering
  \includegraphics[width=0.38\textwidth]{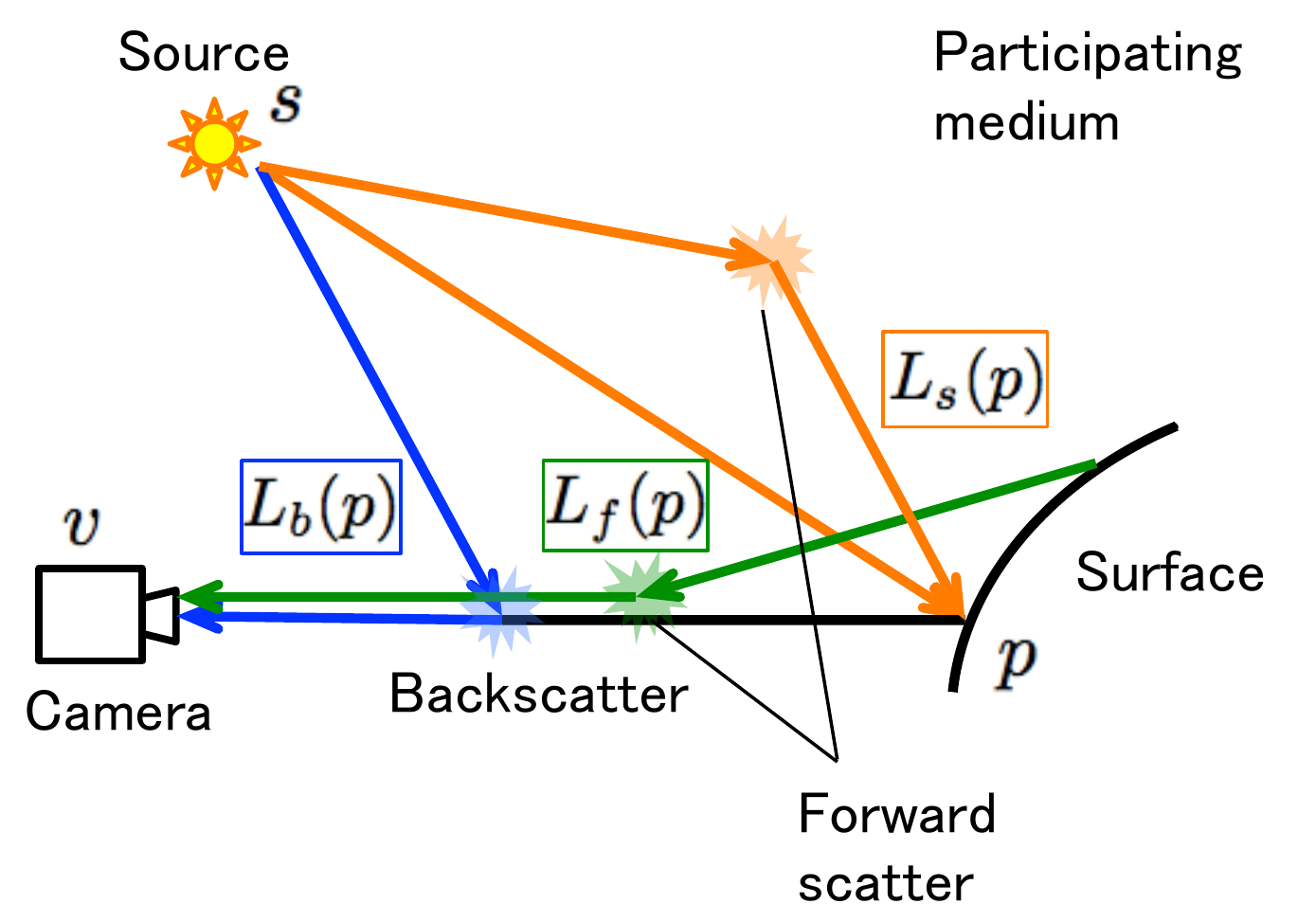}
  \caption{In participating media, the observed irradiance at a camera includes a direct component reflected on a surface, and both backscatter and forward scatter components.}
  \label{fig:light_transport}
\end{figure}

Several methods to reconstruct a 3D shape in participating media using photometric stereo techniques have been proposed \cite{narasimhan05,tsiotsios14,murez17}.
Photometric stereo methods reconstruct surface normals from images captured under different lighting conditions \cite{woodham80}.
Note that backscatter and forward scatter occur in participating media, as shown in Figure \ref{fig:light_transport}; thus, the irradiance observed at a camera includes a direct component reflected on the surface, as well as a backscatter and forward scatter components.
Narasimhan \etal \cite{narasimhan05} modeled single backscattering under a directional light source in participating media and estimated surface normals using a nonlinear optimization technique.
Tsiotsios \etal \cite{tsiotsios14} assumed that backscatter saturates close to the camera when illumination follows the inverse square law, and subtracted the backscatter from the captured image. 

These methods do not consider forward scatter.
Forward scatter depends on the object's shape locally and globally, and 
in highly turbid media such as port water, 3D reconstruction accuracy is affected by forward scatter.
Although Murez \etal \cite{murez17} proposed a photometric stereo technique that considers forward scatter, they assumed that the scene is approximated as a plane, which enables prior calibration.
Therefore, this assumption deteriorates the estimation of normals because forward scatter is intrinsically dependent on the object's shape.

We propose a forward scatter model and implement the model into a photometric stereo framework.
Differing from previous studies \cite{negahdaripour02, murez17}, we compute forward scatter, which depends on the object's shape.
To overcome the mutual dependence between shape and forward scatter, we develop an iterative algorithm that performs a forward scatter removal and 3D shape reconstruction alternately.

In the proposed model, forward scatter is represented by an analytical form of single scattering.
In computer graphics, Monte Carlo and finite element techniques have been used to simulate light scattering in participating media.
Although such techniques provide accurate simulations, realtime rendering is difficult.
Thus, analytical or closed-form solutions have been proposed for efficient computation \cite{sun05, zhou07, pegoraro10}.
For example, Sun \etal \cite{sun05} proposed an analytical single scattering model of backscatter and forward scatter between the source and the surface (source-surface forward scatter) using 2D lookup tables.
Similar to their model, in this study, forward scatter between the surface and the camera (surface-camera forward scatter) is computed using a lookup table. 

Note that surface-camera forward scatter causes image blur.
As mentioned previously, Murez \etal \cite{murez17} assumed that the object is approximated as a plane.
Thus, they modeled forward scatter as a spatially-invariant point spread function under orthogonal projection.
In our proposed model, forward scatter is modeled as spatially-variant kernels because it depends on the object's shape.
Thus, it is impossible to remove forward scatter from the captured image directly because the spatially-variant kernels result in a shift-variant and large-scale dense matrix (Section \ref{sec:approximation_of_matrix}).
To address this problem, we approximate the kernel matrix as a sparse matrix.
We leverage kernel convergence between distant points on the surface for the approximation.

The primary contributions of this paper are summarized as follows:
\begin{itemize}
  \setlength{\parskip}{-3pt}
  \item To the best of our knowledge, this is the first study to strictly model shape-dependent surface-camera forward scatter and we derive its analytical solution.
  \item To remove surface-camera forward scatter, we propose an approximation of the large-scale dense matrix to a sparse matrix.
\end{itemize}

\section{Related work}
\subsection{3D reconstruction in participating media}
In participating media, suspended particles cause light scattering and attenuation that reduce the contrast in captured images.
Nayar \etal \cite{nayar06} proposed direct and global light separation such as interreflection or volumetric scattering using high-frequency patterns.
Treibitz and Schechner \cite{treibitz09} utilized polarization to remove a backscatter component and estimated a depth map from the extracted backscatter.
Kim \etal \cite{kim10} mounted a lenslet array and a diffuser between a camera and a participating medium to estimate blur caused by scattering.
    
Several attempts have been made to design traditional 3D reconstruction techniques for participating media (\eg, structured light \cite{narasimhan05,gu13} and stereo \cite{roser14,negahdaripour14}).
Narasimhan \etal \cite{narasimhan05} proposed a structured light method in participating media, and Gu \etal \cite{gu13} reconstructed the 3D shape of a participating medium using structured light.
Negahdaripour and Sarafraz \cite{negahdaripour14} improved stereo matching in participating media by exploiting the relationship between backscatter and a disparity map.
Recently, Tiang \etal \cite{tiang17} proposed a depth map estimation method using a light field.
Some methods have utilized scattering or attenuation directly for 3D reconstruction.
For example, Hirufuji \etal \cite{hirofuji14} reconstructed specular objects with occlusions using single scattered light, and Inoshita \etal \cite{inoshita12} reconstructed translucent objects directly from volumeric scattering.
Hullin \etal \cite{hullin08} used fluorescence as a participating medium to reconstruct transparent objects, and Asano \etal \cite{asano16} utilized absorption of infrared light to estimate a depth map in underwater scenes.
    
Several photometric stereo methods have been proposed \cite{negahdaripour02, narasimhan05, tsiotsios14, murez17}.
Photometric stereo has several advantages, \eg,
it does not require stereo correspondence and provides pixel-wise detailed shape information even if the target object has a textureless surface.
However, the image formation must be strictly modeled to preserve photometric information.
Narasimhan \etal \cite{narasimhan05} modeled the single scattering of backscatter under a directional light source in participating media, and Tsiotsios \etal \cite{tsiotsios14} demonstrated the saturation of backscatter under the inverse square law, which enabled backscatter removal by subtracting no object image from an input image.
Note that these methods did not consider the effect of forward scatter, which deteriorates the accuracy of 3D reconstruction in highly turbid media.
On the other hand, some methods have considered forward scatter \cite{negahdaripour02, murez17}.
For exmaple, Murez \etal \cite{murez17} approximated the scene as a plane and pre-calibrated the forward scatter component.
Nevertheless, unlike the proposed model, such method do not discuss the relationship between forward scatter and the object's shape.
In this paper, we model shape-dependent surface-camera forward scatter.

\subsection{Analytical solution for single scattering}
In computer graphics, analytical or closed-form solutions for single scattering in participating media have been proposed to overcome computational complexity issues.
Sun \etal \cite{sun05} assumed single and isotropic scattering and used 2D lookup tables to analytically describe backscatter and source-surface forward scatter.
Zhou \etal \cite{zhou07} extended this approach to inhomogeneous single scattering media with respect to backscatter.
Pegoraro \etal \cite{pegoraro10} derived a closed-form solution for single backscattering under a general phase function and light distribution. 
In this study, owing to its simplicity, we use a lookup table similar to that of Sun \etal \cite{sun05}, and we model surface-camera forward scatter analytically. 

\section{Image formation model} \label{sec:image_formation_model}
In this section, we discuss an image formation model in participating media, and provide an analytical form using lookup tables. 
We assume perspective projection, near lighting, and Lambertian objects.
As in many previous studies \cite{narasimhan05, tsiotsios14, murez17, sun05}, multiple scattering is considered to be negligible.

Here, let $L(p)$ be irradiance at a camera when the 3D position $p$ on an object surface is observed. 
In participating media, $L(p)$ is decomposed into a reflected component $L_s(p)$ (Figure \ref{fig:reflected}), a backscatter component $L_b(p)$ (Figure \ref{fig:backscatter}), and a forward scatter component $L_f(p)$ (Figure \ref{fig:forwardscatter}) as follows:
\begin{equation}
L(p) = L_s(p) e^{-cd_{vp}} + L_b(p) + L_f(p).
\label{eq:components}
\end{equation}
Here, parameters $c$ and $d_{vp}$ denote an extinction coefficient and the distance between the camera and position $p$, respectively.
In participating media, light is attenuated exponentially relative to distance.
The extinction coefficient $c$ is the sum of the absorption coefficient $a$ and the scattering coefficient $b$.
\begin{equation}
c = a + b.
\end{equation}

\subsection{Backscatter component}\label{sec:backscatter}
\begin{figure}[t]
  \centering
  \includegraphics[width=0.38\textwidth]{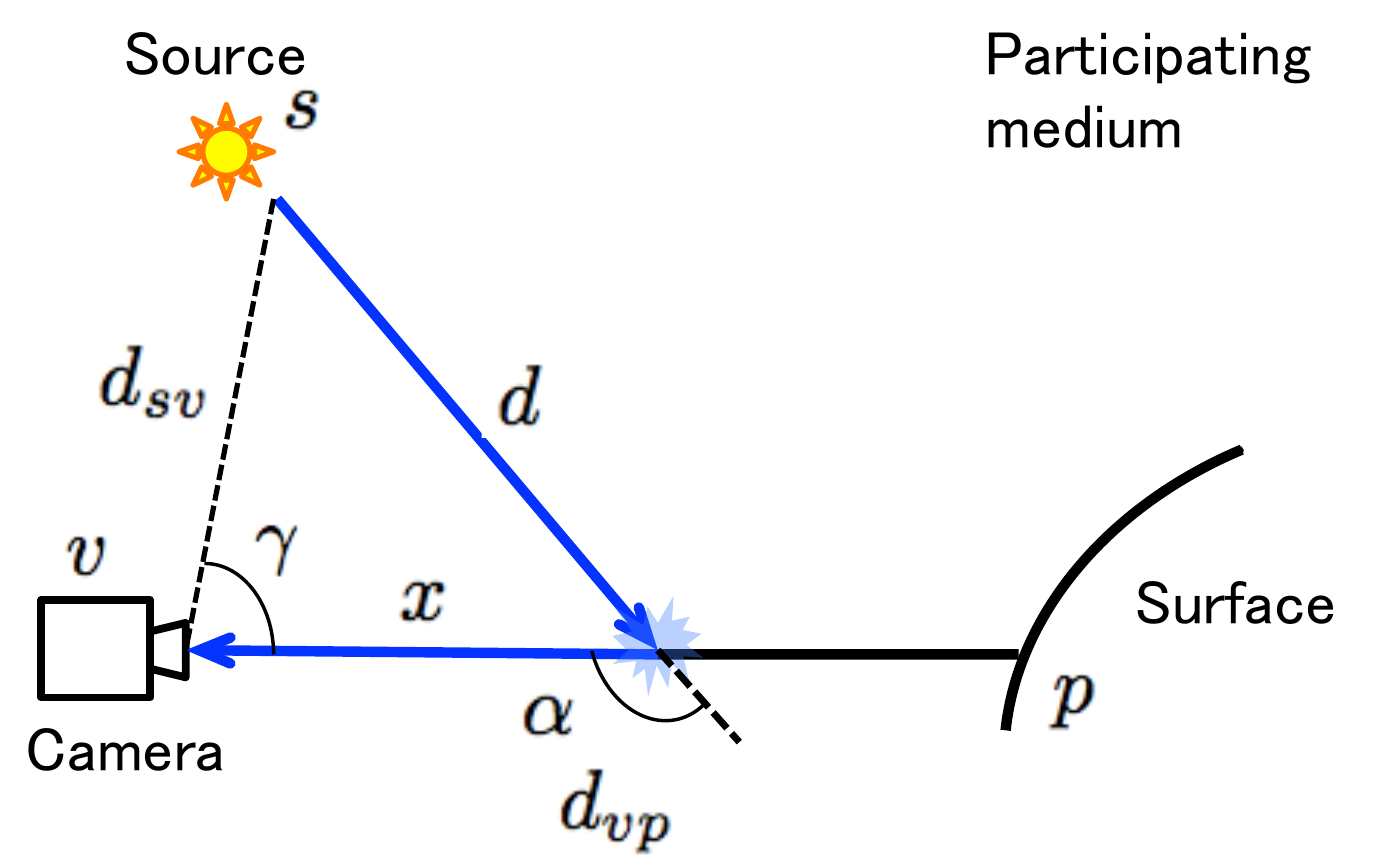}
  \caption{Backscatter component is the sum of scattered light on the viewline without reaching the surface}
  \label{fig:backscatter}
\end{figure}

As shown in Figure \ref{fig:backscatter}, the backscatter component is the sum of scattered light on the viewline without reaching the surface.
Thus, the irradiance of the backscatter component is integral along the line, which is expressed as follows:
\begin{equation}
L_b(p) = \int_0^{d_{vp}}\frac{I_0}{d^2}b{\cal P}(\alpha)e^{-c(x+d)}dx,
\label{eq:backscatter}
\end{equation} 
where $I_0$ denotes the radiant intensity of the source and
${\cal P}(\alpha)$ is a phase function that describes the angular scattering distribution.
Although Equation (\ref{eq:backscatter}) cannot be computed in closed-form, an analytical solution can be acquired using a lookup table.
However, Equation (\ref{eq:backscatter}) depends on $d_{vp}$, $d_{sv}$, $\gamma$, and $c$; thus, the entry of the table is four-dimensional.
Sun \etal \cite{sun05} assumed isotropic scattering (i.e., ${\cal P}(\alpha) = 1/4\pi$) and derived an analytical solution using a 2D lookup table $F(u,v)$:
\begin{flalign}
&L_b(p) = I_0 H_0(T_{sv}, \gamma)& \nonumber \\
&\left[ F(H_1(T_{sv}, \gamma), H_2(T_{vp}, T_{sv}, \gamma)) - F(H_1(T_{sv}, \gamma), \frac{\gamma}{2})\right],&
\label{eq:backscatter_lookup}
\end{flalign}
where $T_{sv}=cd_{sv}$ and $T_{vp} = cd_{vp}$ are optical thickness.
In the following, $T_{xy}$ denotes the product of $c$ and distance $d_{xy}$.
$H_0(T_{sv}, \gamma)$, $H_1(T_{sv}, \gamma)$, and $H_2(T_{vp}, T_{sv}, \gamma)$ are defined as follows:
\begin{eqnarray}
H_0(T_{sv}, \gamma) &=& \frac{bce^{-T_{sv}\cos\gamma}}{2\pi T_{sv}\sin \gamma}, \\
H_1(T_{sv}, \gamma) &=& T_{sv}\sin \gamma, \\
H_2(T_{vp}, T_{sv}, \gamma) &=& \frac{\pi}{4}+\frac{1}{2}\arctan\frac{T_{vp}-T_{sv}\cos \gamma}{T_{sv}\sin \gamma}. 
\end{eqnarray}
$F(u,v)=\int_0^ve^{-u\tan \xi}d\xi$ is a 2D lookup table computed numerically in advance.

As mentioned previously, to remove backscatter, Tsiotsios \etal \cite{tsiotsios14} leveraged backscatter saturation without computing it explicitly.
We also use an image without the target object to remove the backscatter component $L_b(p)$ from the input image. 

\subsection{Reflected component}
\begin{figure}[t]
  \centering
  \includegraphics[width=0.38\textwidth]{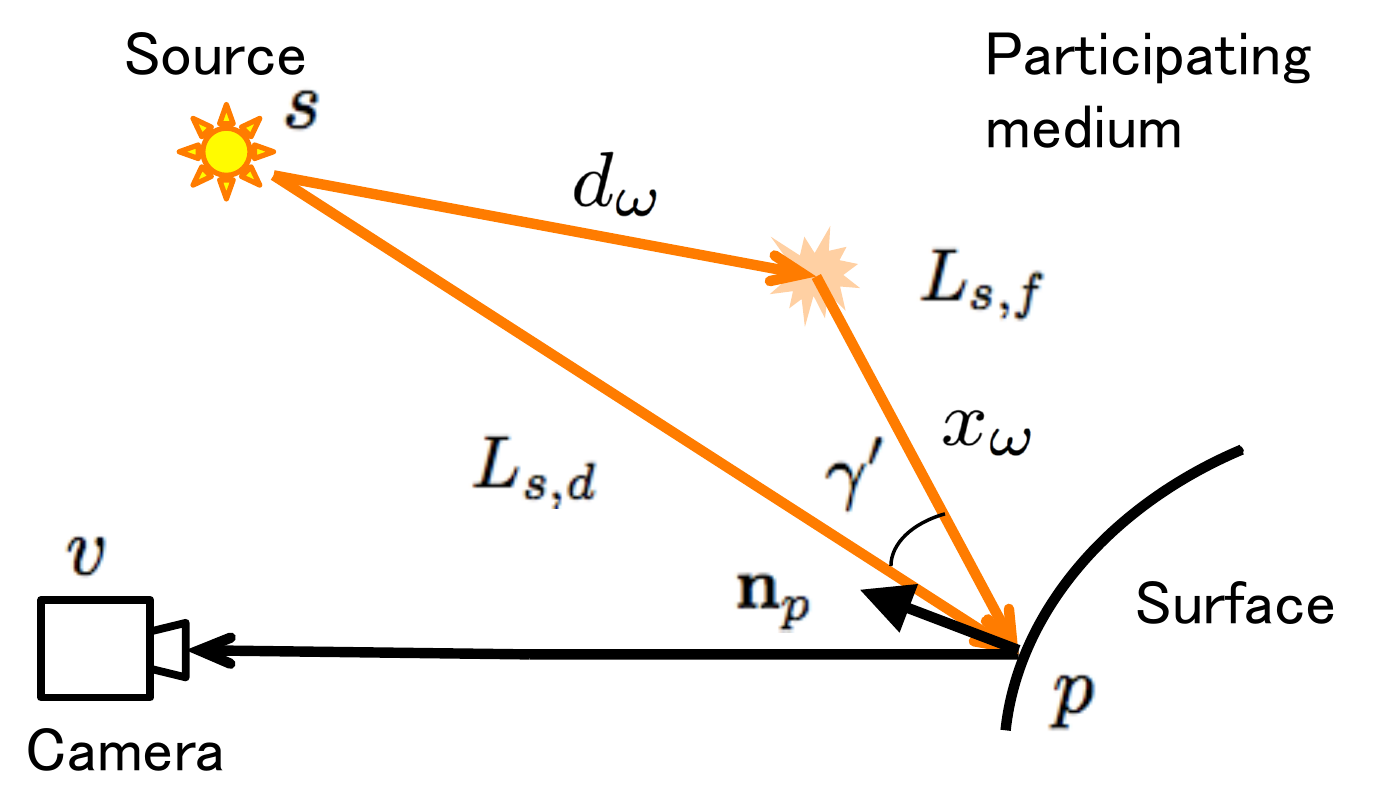}
  \caption{Reflected component is decomposed into $L_{s,d}(p)$ directly reaching the surface from the source and the source-surface forward scatter component $L_{s,f}(p)$.}
  \label{fig:reflected}
\end{figure}

As shown in Figure \ref{fig:reflected}, the reflected component is decomposed into $L_{s,d}(p)$ directly reaching the surface and the source-surface forward scatter component $L_{s,f}(p)$:
\begin{equation}
L_s(p) = L_{s,d}(p) + L_{s,f}(p).
\label{eq:reflected}
\end{equation}
Considering diffuse reflection and attenuation in participating media, $L_{s,d}(p)$ is expressed as follows:
\begin{equation}
L_{s,d}(p) = \frac{I_0}{d_{sp}^2}e^{-T_{sp}}\rho_p \mathbf{n}_p^\top\mathbf{l}_{sp},
\label{eq:directly_reflected}
\end{equation}
where $\rho_p$ is a diffuse albedo at $p$, $\mathbf{n}_p$ is a normal vector, and $\mathbf{l}_{sp}$ is the direction from $p$ to the source.
The source-surface forward scatter component is the integral of scattered light on a hemisphere centered on $p$:
\begin{equation}
L_{s,f}(p) = \int_{\Omega_{2\pi}} L_b(\omega)\rho_p \mathbf{n}_p^\top \mathbf{l_{\omega}}d\omega.
\end{equation}
We define $L_b(\omega)$ as the sum of scattered light from direction $\omega$.
As discussed in Section \ref{sec:backscatter}, Sun \etal \cite{sun05} derived an analytical solution using a 2D lookup table as follows:
\begin{equation}
L_{s,f}(p) = \frac{bcI_0\rho_p}{2\pi T_{sp}}G(T_{sp}, \mathbf{n}_p^\top \mathbf{l}_{sp}),
\label{eq:forwardscatter_surface}
\end{equation}
where $G(T_{sp}, \mathbf{n}_p^\top \mathbf{l}_{sp})$ is a 2D lookup table given as
\begin{flalign}
&G(T_{sp}, \mathbf{n}_p^\top \mathbf{l}_{sp}) = \int_{\Omega_{2\pi}} \frac{e^{-T_{sp}\cos \gamma'}}{\sin \gamma'} & \nonumber \\
&\left[ F(H_1(T_{sp}, \gamma'), \frac{\pi}{2}) - F(H_1(T_{sp}, \gamma'), \frac{\gamma'}{2})\right] \mathbf{n}_p^\top \mathbf{l_{\omega}} d\omega. &
\end{flalign}

\subsection{Surface-camera forward scatter component}
\begin{figure}[t]
  \centering
  \includegraphics[width=0.38\textwidth]{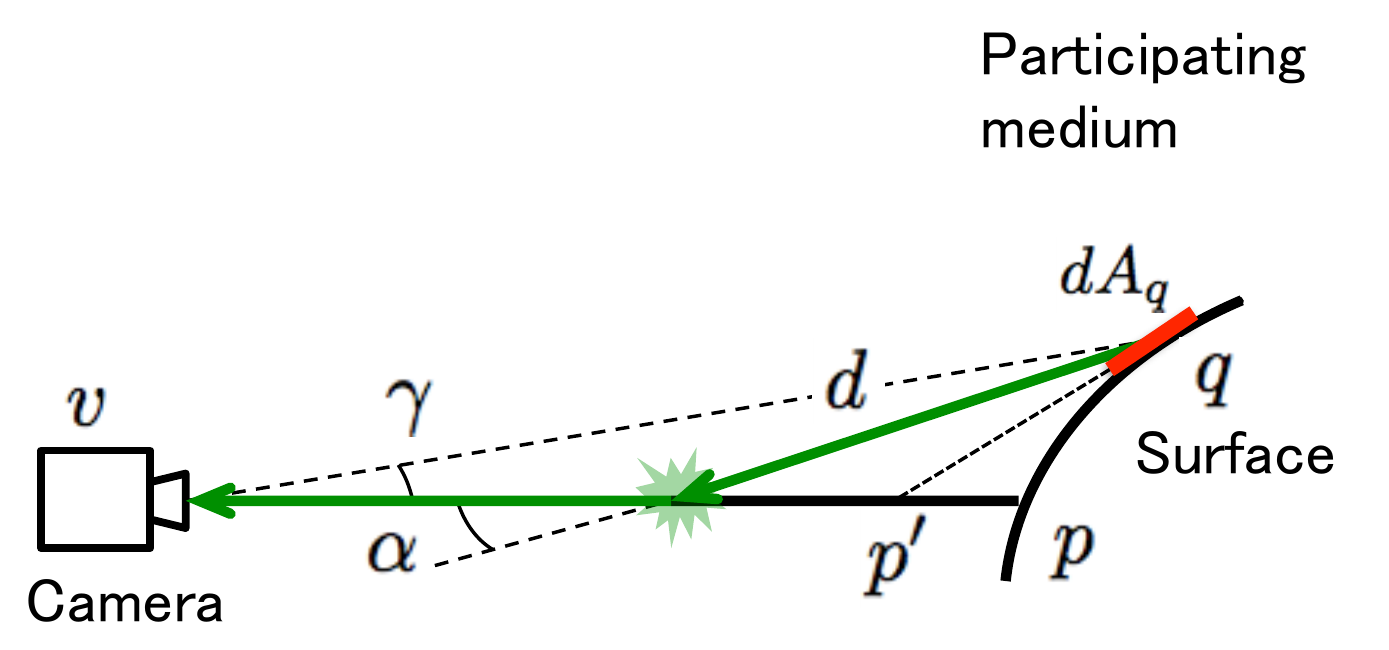}
  \caption{Surface-camera forward scatter component. When we observe surface point $p$ in participating media, the light reflected on point $q$ is scattered on the viewline, and the scattered light is also observed.}
  \label{fig:forwardscatter}
\end{figure}

When we observe surface point $p$ in a participating medium, the light reflected on point $q$ is scattered on the viewline, and the scattered light is also observed as a forward scatter component (Figure \ref{fig:forwardscatter}).
In this paper, we describe this component analytically using a lookup table.

As shown in Figure \ref{fig:forwardscatter}, irradiance at the camera includes reflected light from the small facet centered at $q$.
If we consider this small facet as a virtual light source, similar to Equation (\ref{eq:backscatter}), the irradiance can be expressed as follows:
\begin{equation}
\int_0^{d_{vp'}} \frac{L_s(q)dA_q}{d^2}b{\cal P}(\alpha)e^{-c(x+d)}dx,
\end{equation}
where $dA_q$ is the area of the facet.
At the camera, a discrete point on the surface corresponding to the pixel is observed.
Thus, $L_f(p)$ is the sum of these discrete points:
\begin{equation}
L_f(p) = \sum_{q \neq p}\int_0^{d_{vp'}} \frac{L_s(q)dA_q}{d^2}b{\cal P}(\alpha)e^{-c(x+d)}dx
\end{equation}
Note that the domain of integration $[0,d_{vp'}]$ differs from that of Equation (\ref{eq:backscatter}), i.e., $[0,d_{vp}]$.
We define $p'$ as the intersection point of the viewline and the tangent plane to $q$.
If $d_{vp'} > d_{vp}$, i.e., $p'$ is inside the object, we set $d_{vp'} = d_{vp}$.
If $d_{vp'} < 0$ which means that $p'$ is behind the camera, we set $d_{vp'} = 0$.
Similar to Equation (\ref{eq:backscatter_lookup}), the isotropic scattering assumption yields the following:
\begin{flalign}
&L_f(p) = \sum_{q \neq p}L_s(q)dA_qH_0(T_{vq},\gamma)& \nonumber \\
&\left[ F(H_1(T_{vq}, \gamma), H_2(T_{vp'}, T_{vq}, \gamma)) - F(H_1(T_{vq}, \gamma), \frac{\gamma}{2})\right].&
\label{eq:forwardscatter}
\end{flalign}
This is the analytical expression of the surface-camera forward scatter.
Note that we define the area of the small facet as follows \cite{nayar91}:
\begin{equation}
dA_q = \frac{dI}{\mathbf{v_q}^\top \mathbf{n}_q},
\label{eq:area}
\end{equation}
where $dI$ is the area of the camera pixel and $\mathbf{v_q}$ is the direction from $q$ to the camera.

\section{Photometric stereo considering forward scatter}
To reconstruct the surface normals using photometric stereo, we must deal with both the surface-camera and source-surface forward scatter.
In this section, we first discuss how to remove the surface-camera forward scatter, and then we explain a photometric stereo method that considers the source-surface forward scatter.

\subsection{Approximation of a large-scale dense matrix} \label{sec:approximation_of_matrix}
As mentioned previously, we can remove the backscatter using a previously proposed method \cite{tsiotsios14}.
Here, let $\mathbf{L}' \in \mathbb{R}^N$ be a backscatter subtracted image where $N$ is a number of pixels. Then from Equation (\ref{eq:components}) and (\ref{eq:forwardscatter}), reflected light at the surface $\mathbf{L}_s \in \mathbb{R}^N$ is expressed as follows:
\begin{equation}
\mathbf{L}' = \mathbf{K} \mathbf{L}_s,
\label{eq:linear}
\end{equation}
where $\mathbf{K}$ is an $N \times N$ dense matrix.
$\mathbf{K}$ is a large-scale dense matrix whose elements are given by
\begin{flalign}
&K_{pq} = \nonumber\\
&\left\{
\begin{array}{l}
e^{-T_{vp}}\;\;\;\;\;\;(p = q)\\
dA_qH_0(T_{vq},\gamma)\left[ F(H_1(T_{vq}, \gamma), H_2(T_{vp'}, T_{vq}, \gamma)) \right.\\
\left. - F(H_1(T_{vq}, \gamma), \frac{\gamma}{2}) \right] \;\;\;\;\;\; (p \neq q) .
\end{array}
\right.
\label{eq:kernel}
\end{flalign}

Our model is similar to that of Murez \etal \cite{murez17}.
However, our model is different in that each row of $\mathbf{K}$ is spatially-variant because we compute the forward scatter considering the object's shape.
In the model presented by Murez \etal \cite{murez17}, the plane approximation of the scene under orthogonal projection yields a spatially-invariant point spread function.
Our spatially-variant kernel matrix makes it impossible to solve Equation (\ref{eq:linear}) directly.

\begin{figure}[t]
  \begin{minipage}{0.49\hsize}
    \centering
      \includegraphics[scale=0.3]{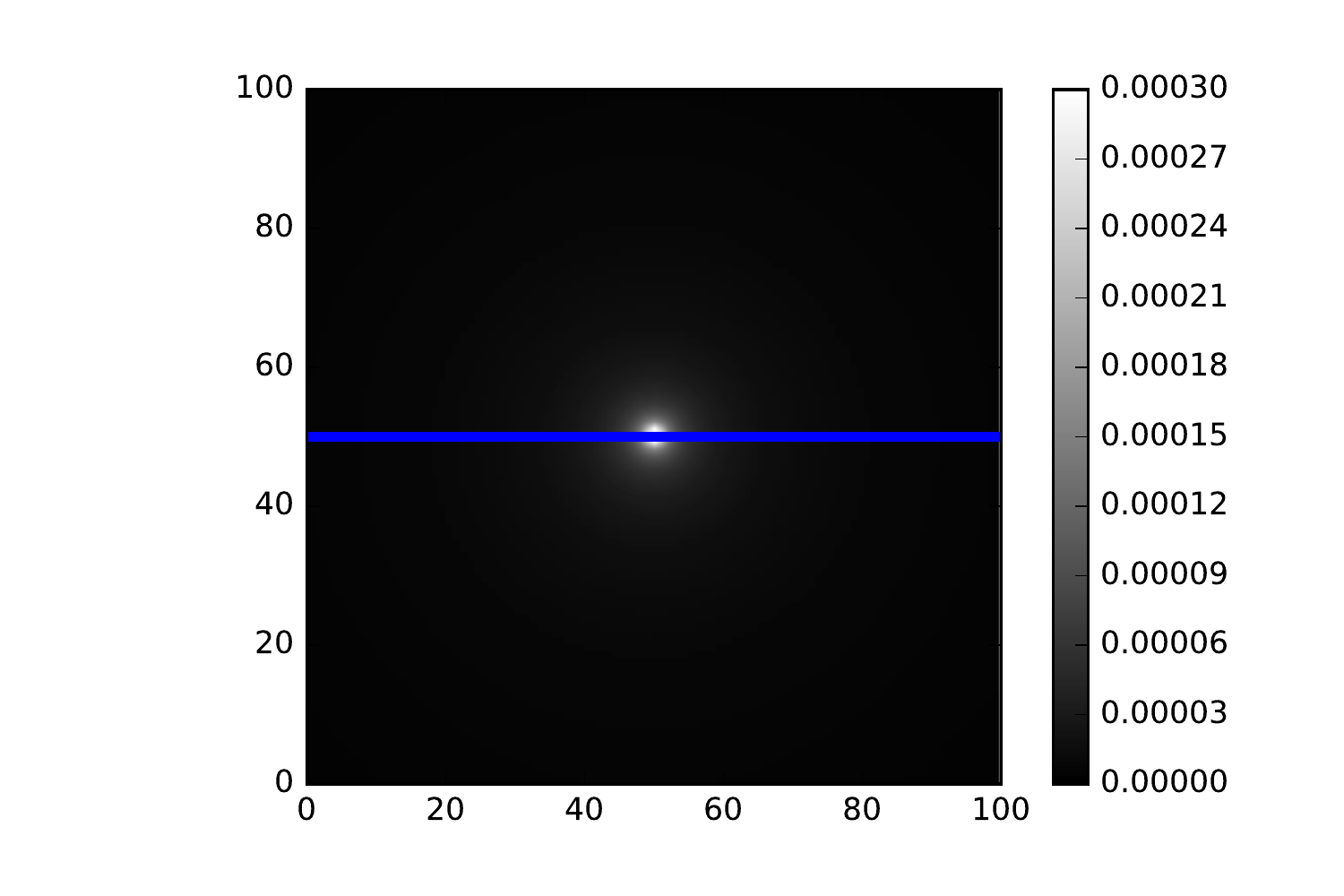}
    \subcaption{}
    \label{fig:2dplot}
  \end{minipage}
  \begin{minipage}{0.49\hsize}
    \centering
      \includegraphics[scale=0.3]{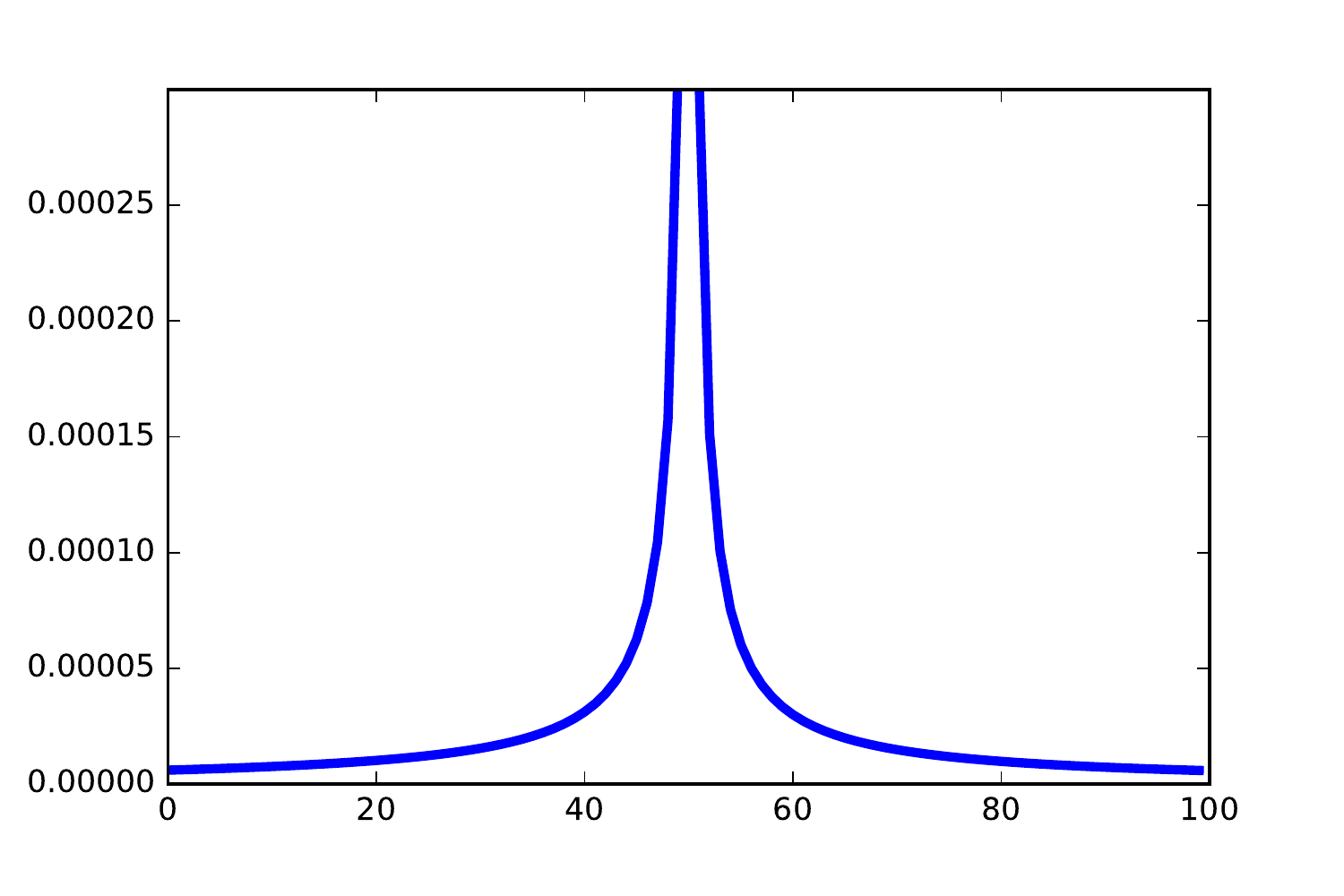}
    \subcaption{}
    \label{fig:profile}
  \end{minipage}
  \caption{(a) 2D visualization of a row of $\mathbf{K}$ when we observe a plane; (b) profile of the blue line in (a). These figures show that the effect between two points converges to a very small value as the distance of the points increases.}
  \label{fig:convergence}
\end{figure}

To overcome this problem, we propose an approximation of a large-scale dense matrix $\mathbf{K}$ as a sparse matrix.
Figure \ref{fig:convergence} (a) shows a row of $\mathbf{K}$ reshaped in a 2D when we observe a plane in a participating medium.
This shows how the observed irradiance of the center of the plane is affected by other points.
Figure \ref{fig:convergence} (b) shows the profile of the blue line in Figure \ref{fig:convergence} (a).
From these figures, we observe that the effect between two points converges to a very small value as the distance of the points increases; however, it does not converge to zero. 
Here, we assume that the value of $K_{pq}$ converges to $\epsilon\;(0 < \epsilon \ll 1)$ in the neighboring set $S(p)$ centered at $p$, and we obtain the following approximation:
\begin{flalign}
L'(p) &= \sum_{q} K_{pq} L_s(q)&  \\
&\approx \sum_{q \in S(p)} K_{pq} L_s(q) + \sum_{q \notin S(p)} \epsilon L_s(q)& \label{eq:before_approx}\\
&\approx \sum_{q \in S(p)} K_{pq} L_s(q) + C, \label{eq:after_approx}&
\end{flalign}
where $C = \sum_{q} \epsilon L_s(q)$ and we use $\sum_{q \in S(p)} \epsilon L_s(q) \approx 0$ from Equation (\ref{eq:before_approx}) to (\ref{eq:after_approx}).
Then, we define a sparse matrix $\hat{\mathbf{K}}$ as follows:
\begin{equation}
\hat{K}_{pq} = \left\{
\begin{array}{l}
K_{pq} \;\;\; (q \in S(p)) \\
0 \;\;\; (q \notin S(p)).
\end{array}
\right.
\end{equation}
This yields the following linear system:
\begin{eqnarray}
\left[
\begin{array}{c}
\mathbf{L}'\\
0 \\
\end{array}
\right]
=
\left[
\begin{array}{cccc}
&&&1 \\
&\hat{\mathbf{K}}&&\vdots \\
&&&1 \\
\epsilon&\cdots&\epsilon&-1 \\
\end{array}
\right]
\left[
\begin{array}{c}
\mathbf{L}_s \\
C \\
\end{array}
\right].
\label{eq:sparse_linear}
\end{eqnarray}
We solve this linear system using BiCG stabilization \cite{vorst92} to remove surface-camera forward scatter.
We define $S(p)$ as the set of 3D points captured in a $r \times r$ region centered at the observed pixel $p$.
Note that the size of the kernel support $r$ should be set manually. In our experiments, $r=61$ to $r=81$ gave efficient results.
To avoid computation of all the elemetns of $\mathbf{K}$, we approximated the convergence value $\epsilon$ as follows:
\begin{equation}
\epsilon = \min_{p,q} \left\{ K_{pq} \mid q \in S(p) \right\}
\end{equation}

\subsection{Photometric stereo}
After removing the backscatter and surface-camera forward scatter, we can obtain the reflected components $\mathbf{L}_s$.
We reconstruct the surface normals by applying photometric stereo to $\mathbf{L}_s$.
From Equations (\ref{eq:reflected}), (\ref{eq:directly_reflected}) and (\ref{eq:forwardscatter_surface}), $L_s(p)$ is given as follows:
\begin{equation}
L_s(p) = \frac{I_0}{d_{sp}^2}e^{-T_{sp}}\rho_p (\mathbf{n}_p^\top\mathbf{l}_{sp}) + \frac{bcI_0\rho_p}{2\pi T_{sp}}G(T_{sp}, \mathbf{n}_p^\top \mathbf{l}_{sp}).
\label{eq:surface_rewrited}
\end{equation}

\begin{figure}[t]
  \begin{minipage}{0.49\hsize}
    \centering
      \includegraphics[scale=0.3]{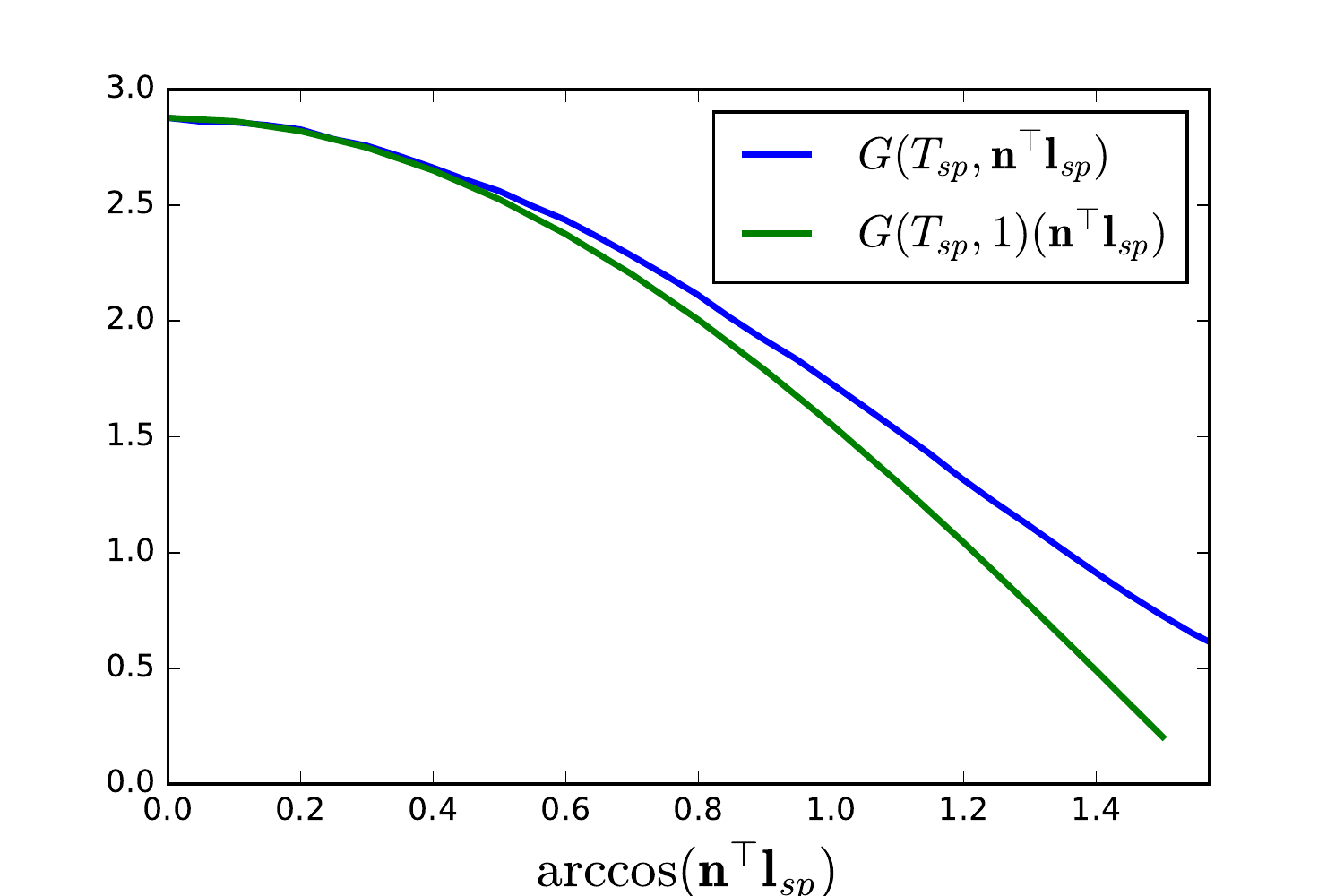}
    \subcaption{}
    \label{fig:g_approx_.6}
  \end{minipage}
  \begin{minipage}{0.49\hsize}
    \centering
      \includegraphics[scale=0.3]{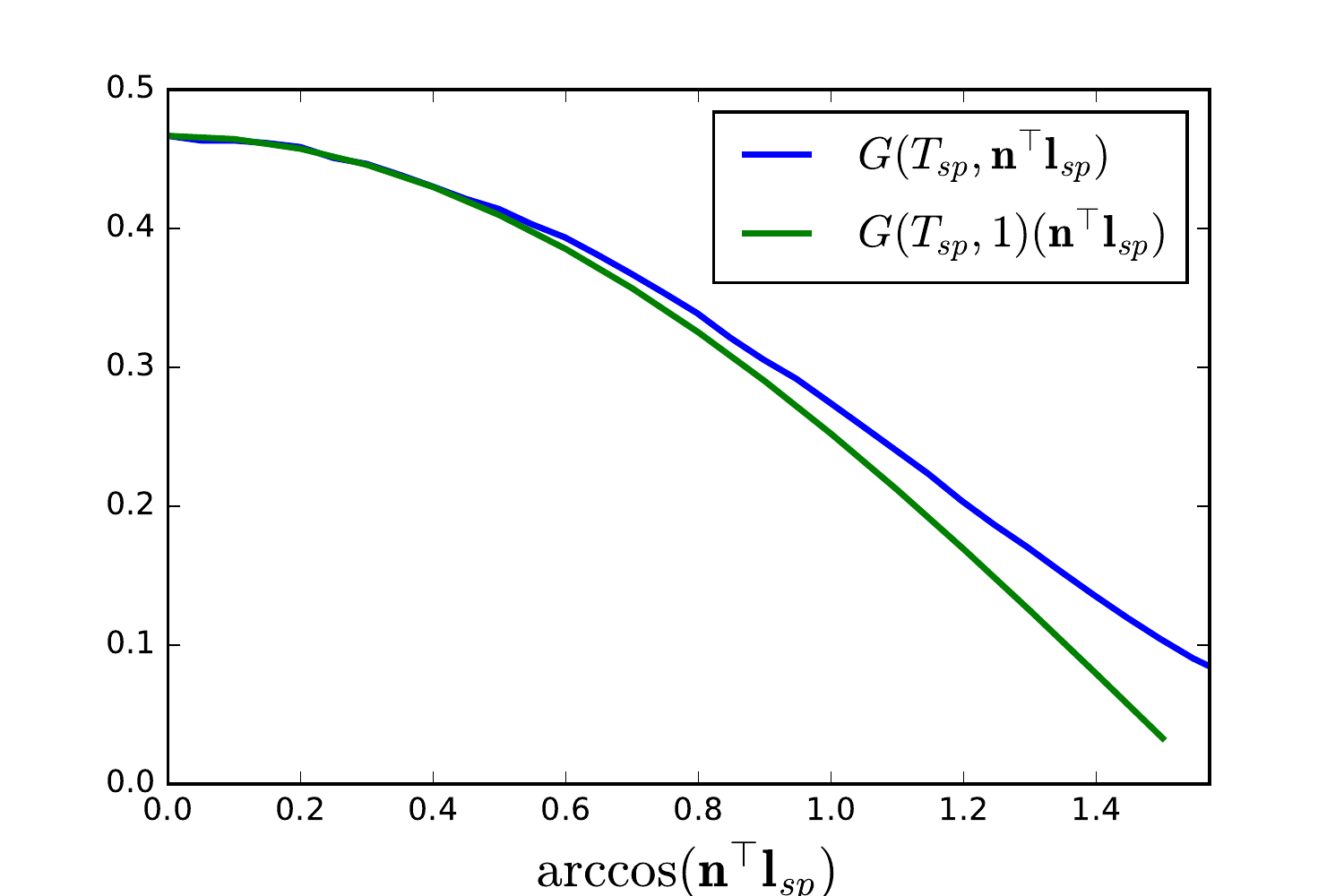}
    \subcaption{}
    \label{fig:g_approx_2}
  \end{minipage}
  \caption{$G(T_{sp}, \mathbf{n}_p^\top \mathbf{l}_{sp})$ (blue line) and $G(T_{sp},1)(\mathbf{n}_p^\top \mathbf{l}_{sp})$ (green line) when (a) $T_{sp} = 0.6$ and (b) $T_{sp} = 2$. Although the error increases as $\arccos (\mathbf{n}_p^\top \mathbf{l}_{sp})$ increases, these graphs validate the approximation $G(T_{sp}, \mathbf{n}_p^\top \mathbf{l}_{sp}) \approx G(T_{sp},1)(\mathbf{n}_p^\top \mathbf{l}_{sp})$.
  }
  \label{fig:g_approx}
\end{figure}

Note that this equation is not linear with respect to the normal due to the source-surface forward scatter.
We want to apply photometric stereo directly to the equation; therefore, we use the following approximation of table $G(T_{sp}, \mathbf{n}_p^\top \mathbf{l}_{sp})$:
\begin{equation}
G(T_{sp}, \mathbf{n}_p^\top \mathbf{l}_{sp}) \approx G(T_{sp},1)(\mathbf{n}_p^\top \mathbf{l}_{sp}).
\end{equation}
In Figure \ref{fig:g_approx}, we plot $G(T_{sp}, \mathbf{n}_p^\top \mathbf{l}_{sp})$ and $G(T_{sp},1)(\mathbf{n}_p^\top \mathbf{l}_{sp})$ when $T_{sp} = 0.6$ and $T_{sp} = 2$.
In each figure, the blue line represents $G(T_{sp}, \mathbf{n}_p^\top \mathbf{l}_{sp})$ and the green line represents $G(T_{sp},1)(\mathbf{n}_p^\top \mathbf{l}_{sp})$.
Although the error increases as $\arccos (\mathbf{n}_p^\top \mathbf{l}_{sp})$ increases, these graphs validate this approximation.
Therefore, we can obtain the following:
\begin{equation}
L_s(p) \approx \rho_p I_0 \left( \frac{e^{-T_{sp}}}{d_{sp}^2} + \frac{bc}{2\pi T_{sp}}G(T_{sp},1)  \right) \mathbf{n}_p^\top \mathbf{l}_{sp}.
\label{eq:photometric_stereo}
\end{equation}
This is a linear equation about normal $\mathbf{n}_p$; hence we apply photometric stereo to this equation.

\subsection{Implementation}
In this section, we explain our overall algorithm.
Note that the kernel of equation \ref{eq:kernel} is only defined on the object's surface; thus, we input a mask image and perform the proposed method on only the object region.
Backscatter is removed using a previously proposed method \cite{tsiotsios14}; however, the resulting image contains high-frequency noise due to SNR degradation.
Therefore, we apply a $3 \times 3$ median filter after removing the backscatter to reduce this high-frequency noise.
We used Poisson solver \cite{agrawal06} which is extended to perspective projection \cite{papadhimitri13} for normal integration to reconstruct the shape.
The proposed algorithm is described as follows:
\begin{enumerate}
  \setlength{\parskip}{-3pt}
  \item Input images and a mask. 
  \\Initialize the shape and normals.
  \item Remove backscatter \cite{tsiotsios14} and apply a median filter to the resulting images.
  \item Remove forward scatter between the object and the camera (Equation (\ref{eq:sparse_linear})).
  \item Reconstruct the normals using Equation (\ref{eq:photometric_stereo}). 
  \item Integrate the normals and update them from the reconstructed shape.
  \item Repeat steps 3--5 until convergence.
\end{enumerate}

\section{Experiments}
\subsection{Experiments with synthesized data}
\begin{figure*}[t]
\centering
  \begin{minipage}{0.14\textwidth}
    \centering
      \includegraphics[scale=0.25]{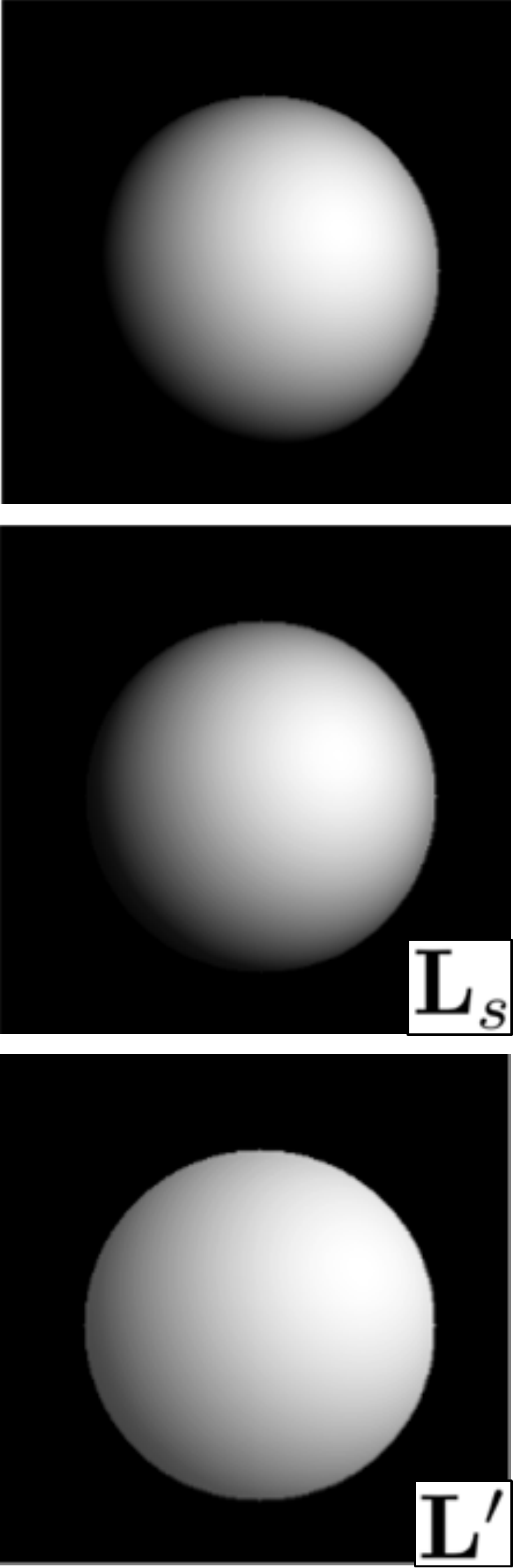}
    \subcaption{}
  \end{minipage}
  \begin{minipage}{0.14\textwidth}
    \centering
      \includegraphics[scale=0.25]{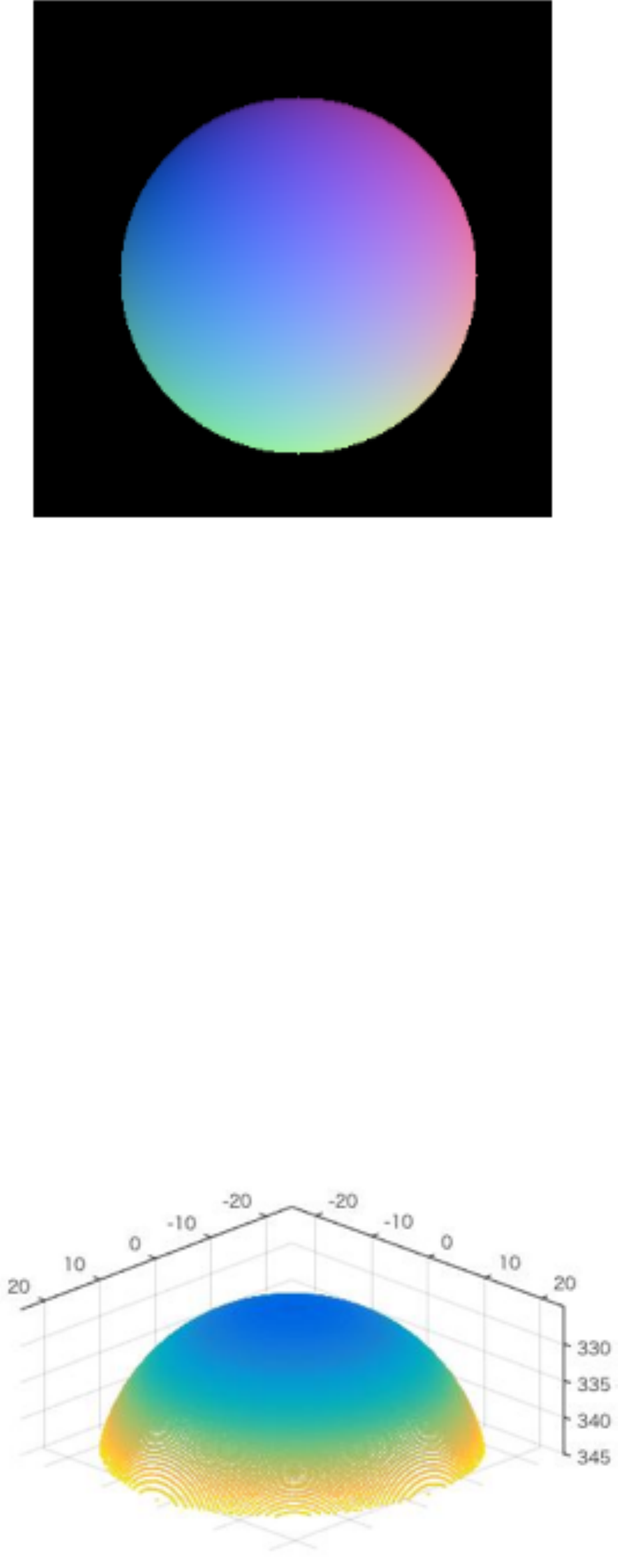}
    \subcaption{}
  \end{minipage}
  \begin{minipage}{0.7\textwidth}
    \centering
      \includegraphics[scale=0.25]{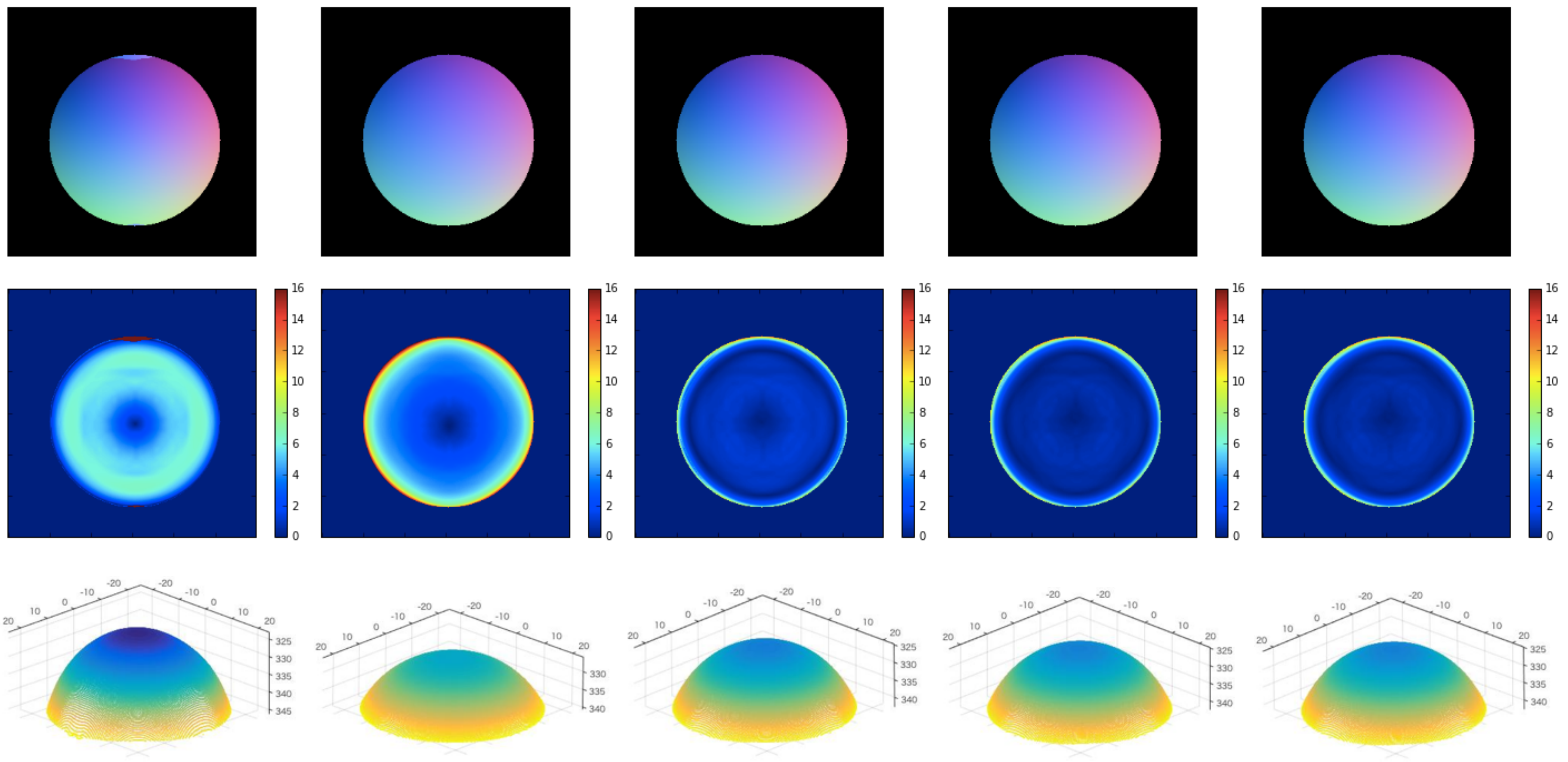}
    \subcaption{}
  \end{minipage}
  \caption{Results of the synthesized data. (a) examples of synthesized data. (b) ground truth. (c) output of the each iteration from left to right. 
  (top row) output normals, (middle row) error map of angles, (bottom row) reconstructed shape.}
  \label{fig:simulation_sphere_exp}
\end{figure*}

We first describe experiments with synthesized data. 
We generated 8 synthsized images with a 3D model of a sphere using our scattering model in Section \ref{sec:image_formation_model}.
The scattering property was assumed to be isotoropic and the parameters were set as $b = c = 5.0 \times 10^{-3}$.
We show the examples of the synthesized images in Figure \ref{fig:simulation_sphere_exp} (a), where an image without a participating medium, a reflected component $\mathbf{L}_s$, and a backscatter subtracted image $\mathbf{L}'$ from top to bottom.
In the experimetns, the kernel support was set as $r=81$.
The shape was initialized as a plane.

The results are shown in Figure \ref{fig:simulation_sphere_exp} (b) (c) and Table \ref{tab:simulation_sphere_exp}.
Figure \ref{fig:simulation_sphere_exp} (b) shows the ground truth and (c) shows the output of each iteration from left ro right.
The top row shows the normals map, the middle row shows the angular error of the output, and the bottom row shows the reconstructed shapes.
Table \ref{tab:simulation_sphere_exp} shows the mean angular error of each output. 
GT in Table \ref{tab:simulation_sphere_exp} denotes the error when we removed scattering effects with the ground truth shape and reconstructed the 3D shape inversely.
As shown in Figure \ref{fig:simulation_sphere_exp}, the shape converged while oscillating in height.
This convergence was also seen in the experiments with the real data (Figure \ref{fig:sphere_expt}).

\subsection{Experimental environment}
\begin{figure}[t]
  \begin{minipage}{0.66\hsize}
    \centering
      \includegraphics[scale=0.55]{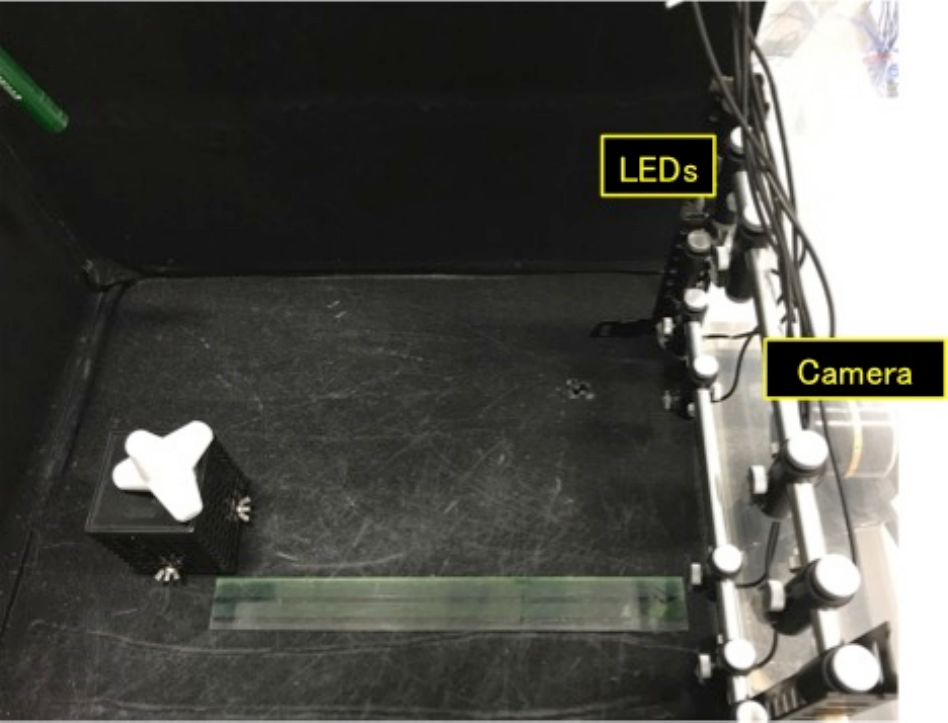}
    \subcaption{}
  \end{minipage}
  \begin{minipage}{0.33\hsize}
    \centering
      \includegraphics[scale=0.55]{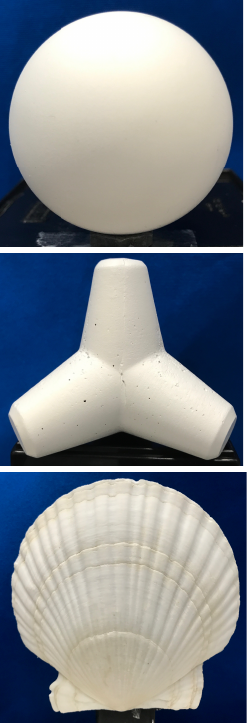}
    \subcaption{}
  \end{minipage}
  \caption{(a) Experimental environment. (b) Target objects.}
  \label{fig:environment_and_objects}
\end{figure}

We also evaluated the proposed methods using real captured data.
The experimental environment is shown in Figure \ref{fig:environment_and_objects} (a).
We used a 60-cm cubic tank and placed a target object in the tank.
We used diluted milk as a participating medium.
The medium parameters were set with reference to the literature \cite{narasimhan06}.
A ViewPLUS Xviii 18-bit linear camera was mounted in close contact with the tank, and eight LEDs were mounted around the camera.
The input images were captured at an exposure of 33 ms.
We captured 60 images under the same condition, and these images were averaged to make input images robust to noise caused by the imaging system; thus, eight averaged images were input to the proposed method.

The camera was calibrated using the method presented in the literature \cite{zhang00}.
To consider refraction on the wall of the tank, calibration was performed when the tank was full of water. 
The locations of the LEDs were measured manually, and each radiant intensity $I_0$ was calibrated using a white Lambertian sphere.

The target objects are shown in Figure \ref{fig:environment_and_objects} (b) ({\it sphere}, {\it tetrapod}, and {\it shell}).

\begin{table}[t]
  \centering
    \begin{tabular}{c|rrrrr|r} \hline
      & 1 & 2 & 3 & 4 & 5 & GT\\ \hline 
      Error (deg.)& 5.20 & 4.65 & 1.43 & 1.29 & 1.29 & 1.30\\ \hline
    \end{tabular}
    \caption{Mean angular error of the output of the each iteration with synthesized data}
    \label{tab:simulation_sphere_exp}
\end{table}

\subsection{Comparison wtih the backscatter-only modeling}
We compared the proposed method with a previously proposed method \cite{tsiotsios14} that models only backscatter.
In each experiment, we initialized the target object as a plane for the iteration, and the kernel support was set as $r=61$.

\begin{figure*}[t]
\centering
  \begin{minipage}[c]{0.14\hsize}
    \centering
      \includegraphics[scale=0.42]{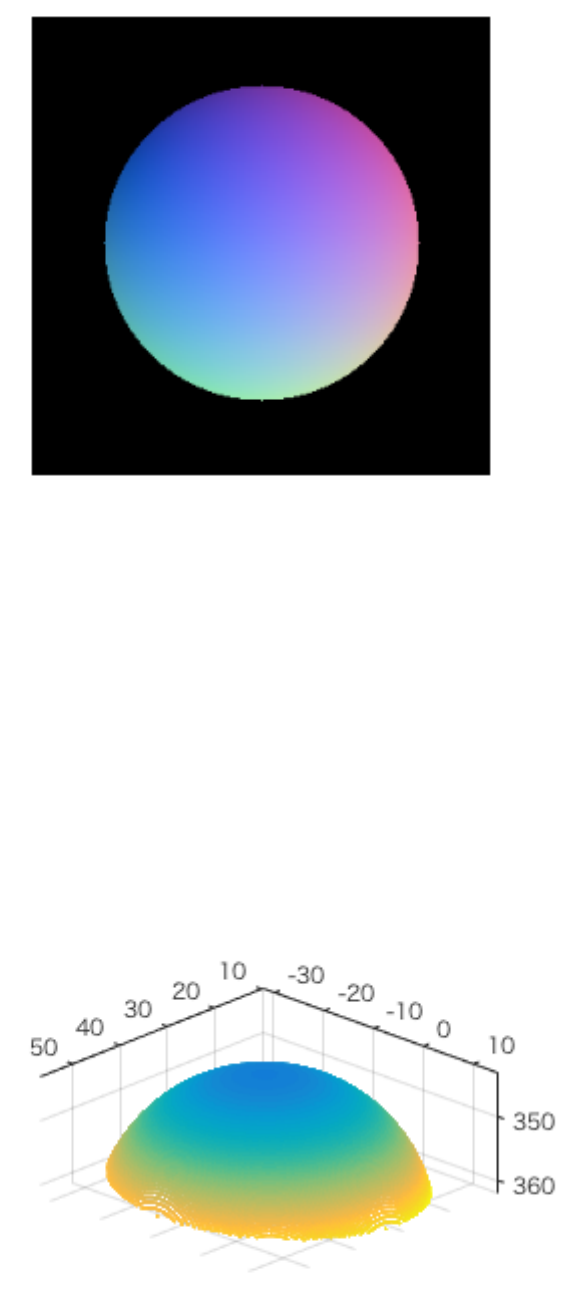}
    \subcaption{}
    \label{fig:sphere_o}
  \end{minipage}
  \begin{minipage}[c]{0.14\hsize}
    \centering
      \includegraphics[scale=0.42]{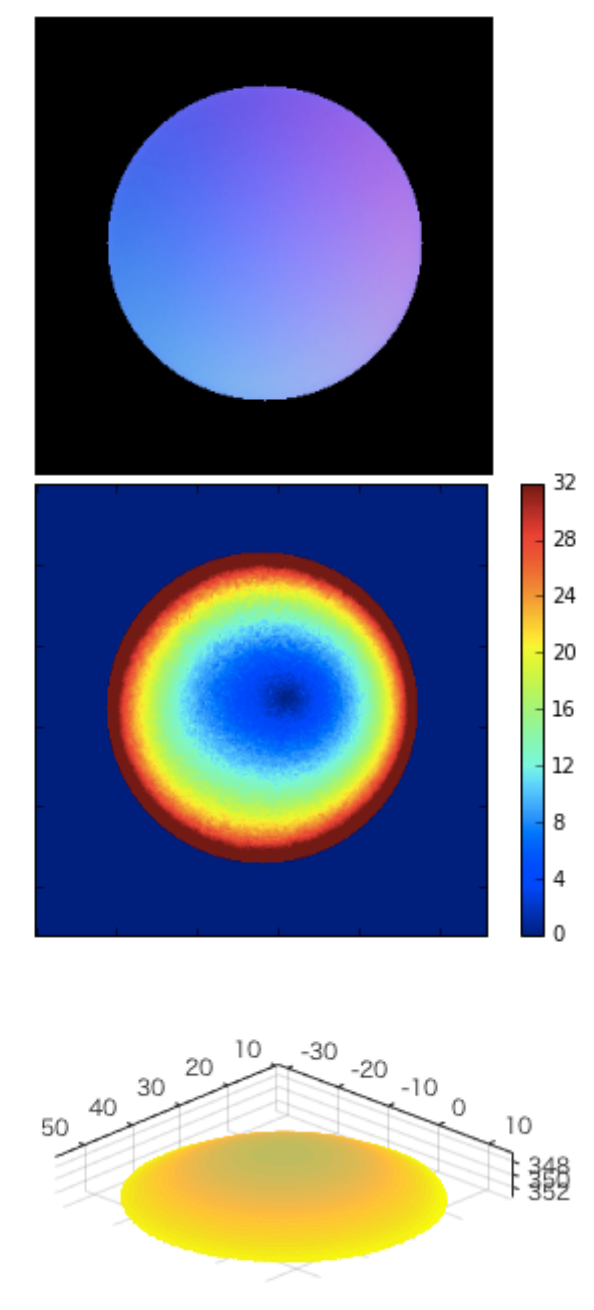}
    \subcaption{}
    \label{fig:tetra_o}
  \end{minipage}
  \begin{minipage}[c]{0.7\hsize}
    \centering
      \includegraphics[scale=0.42]{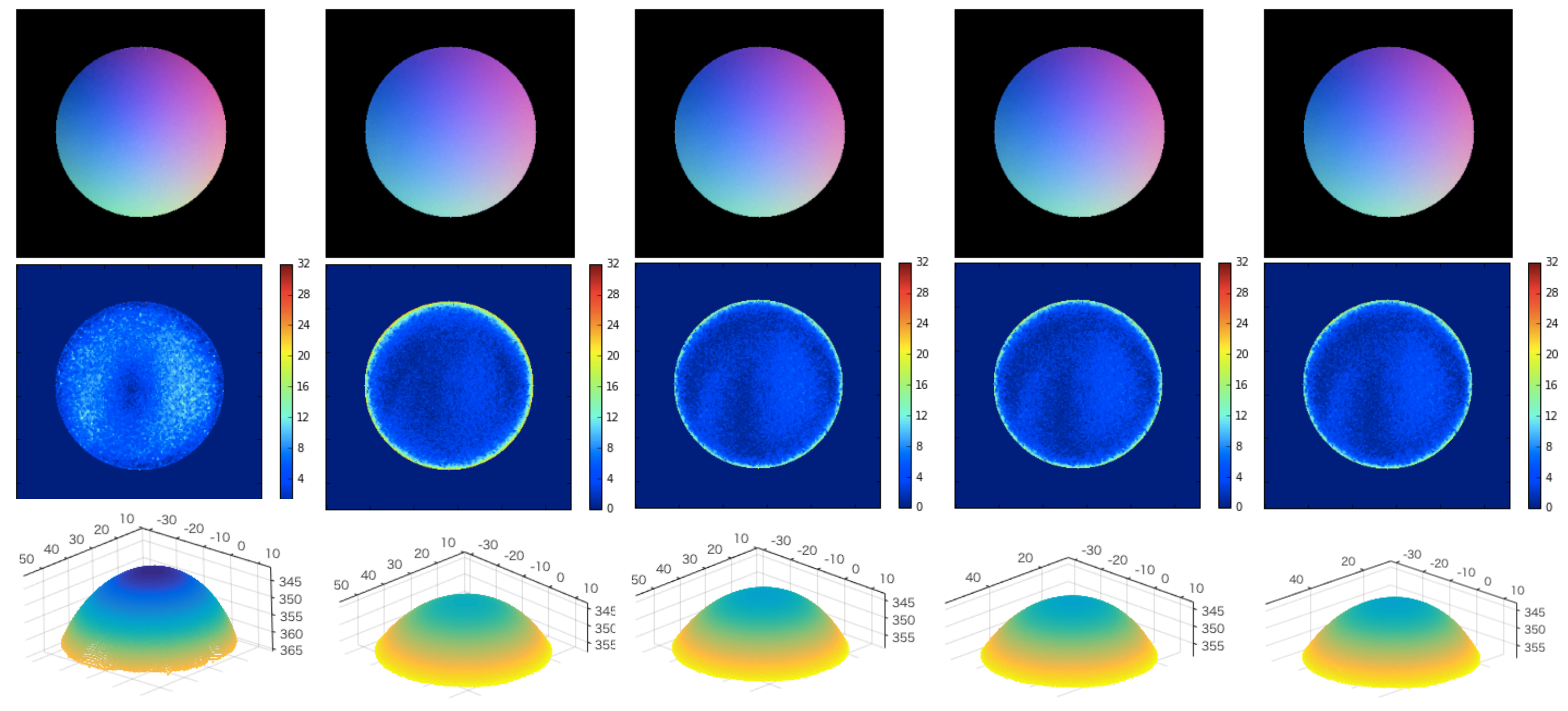}
    \subcaption{}
    \label{fig:shell_o}
  \end{minipage}
  \caption{Results of {\it sphere}; (a) ground truth, (b) result of \cite{tsiotsios14}, and (c) proposed method; 
  (top row) output normals, (middle row) error map of angles, (bottom row) reconstructed shape; 
  }
  \label{fig:sphere_expt}
\end{figure*}

First, we evaluated the proposed method quantitatively using {\it sphere}.
In this experiment, we placed 120 L of water and 30 mL of milk in the tank.
Figure \ref{fig:degraded} (b) shows one of the input images.
The results are given in Figure \ref{fig:sphere_expt}, where (a) shows the ground truth, (b) shows the result of the backscatter-only modeling \cite{tsiotsios14}, and (c) shows the result of the proposed method. 
These experimental results demonstrate that the proposed method can reconstruct the object's shape in highly turbid media, in which the method that does not consider forward scatter fails.
Table \ref{tab:sphere_expt} shows the mean angular error of the results of the backscatter-only modeling \cite{tsiotsios14} and the output of each iteration of the proposed method.
As can be seen, the error reaches convergence during a few iterations.

\begin{figure}[t]
\centering
    \begin{minipage}{0.25\hsize}
      \centering
      \includegraphics[scale=0.3]{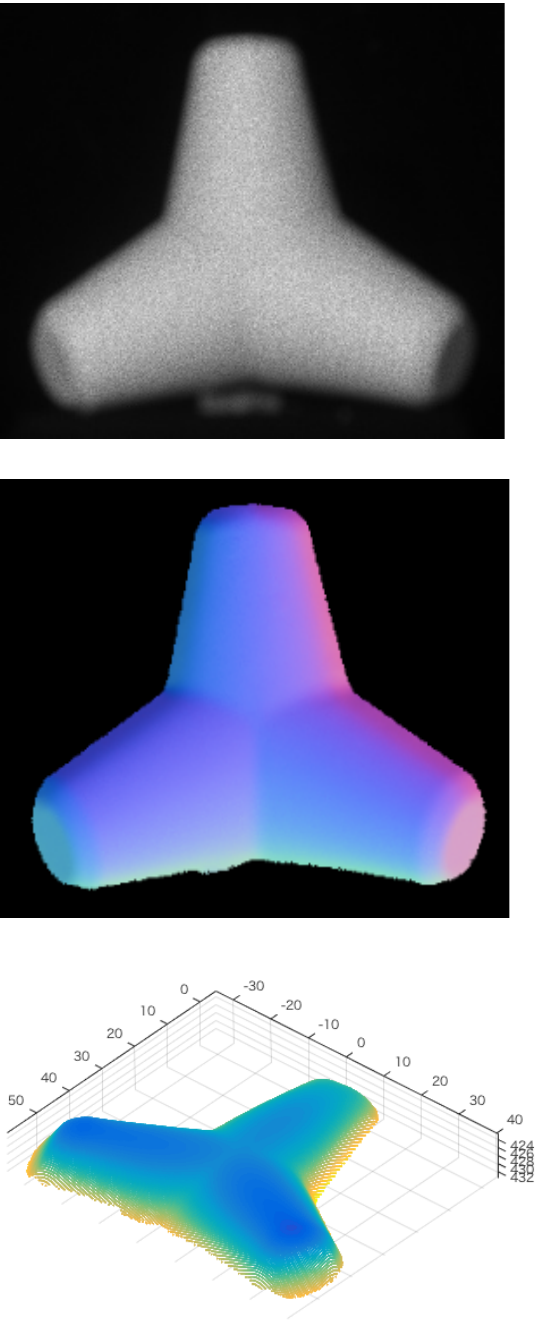}
      \subcaption{}
    \end{minipage}
    \begin{minipage}{0.74\hsize}
      \centering
      \includegraphics[scale=0.3]{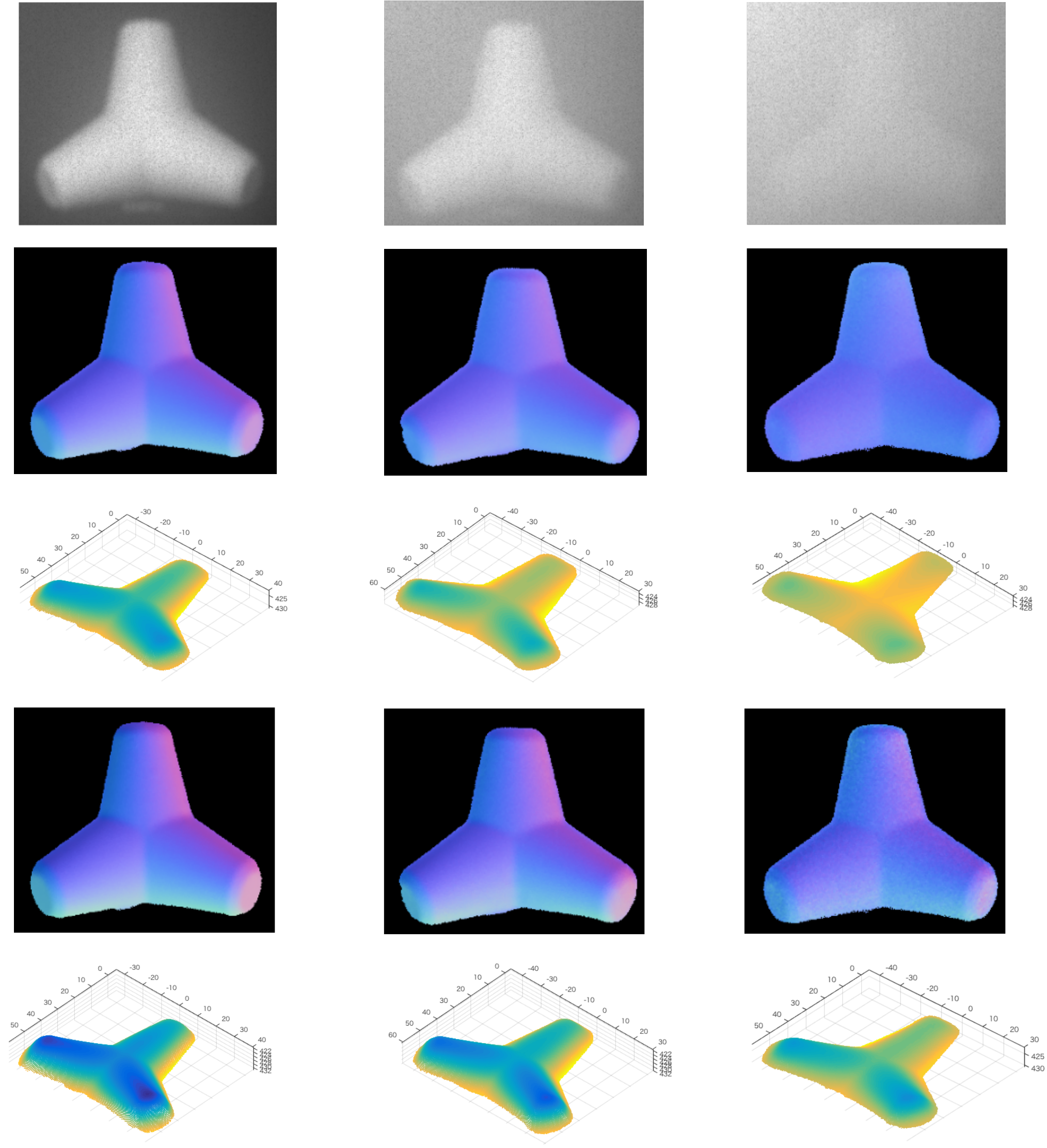}
      \subcaption{}
    \end{minipage}
    \caption{Results of {\it tetrapod}; (a) reconstruction in clear water and (b) results of \cite{tsiotsios14} (second and third rows) and the proposed method (fourth and fifth rows). The top row is one of the input images. The concentration of the participating medium increases from left to right. 
    }
    \label{fig:tetra_expt}
\end{figure}

\begin{figure}[t]
    \begin{minipage}{0.25\hsize}
      \centering
      \includegraphics[scale=0.3]{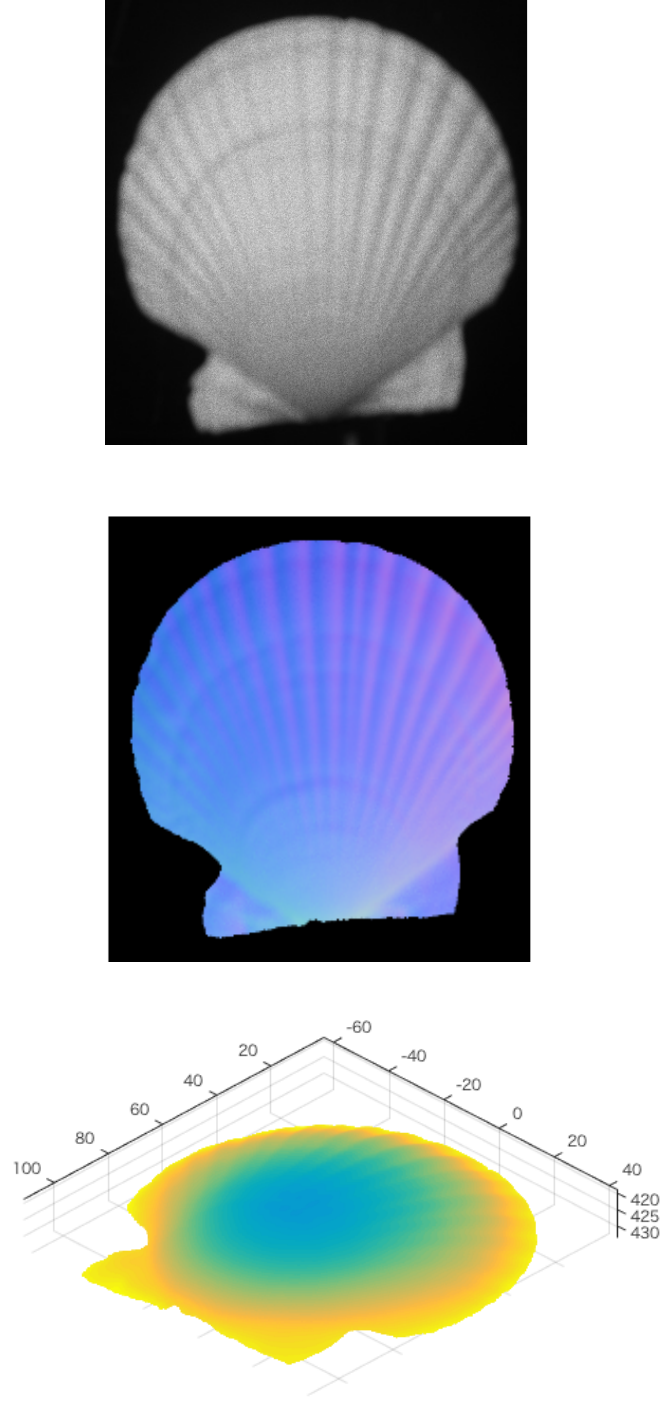}
      \subcaption{}
    \end{minipage}
    \begin{minipage}{0.74\hsize}
      \centering
      \includegraphics[scale=0.3]{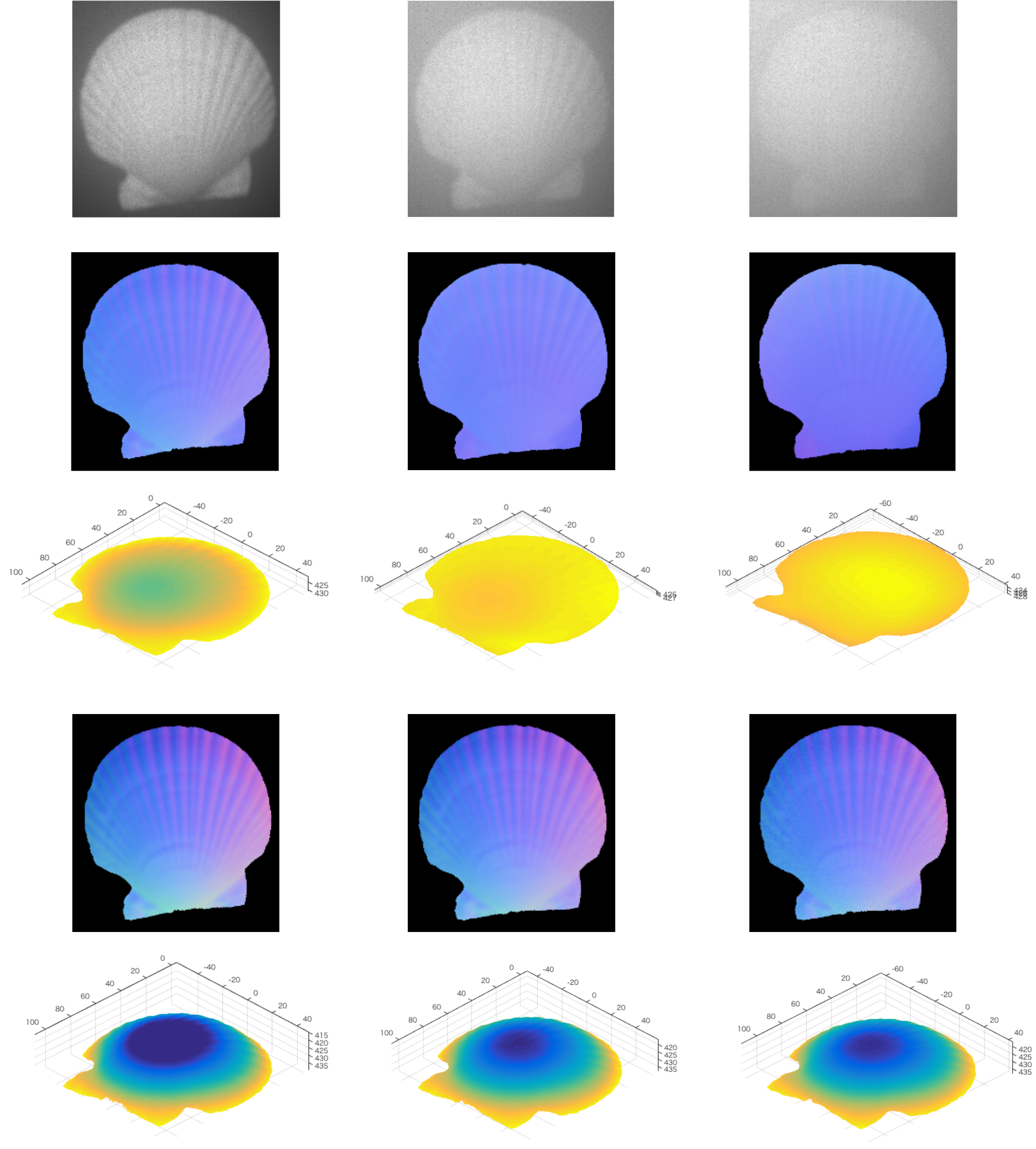}
      \subcaption{}
    \end{minipage}
    \caption{Results of {\it shell}; details are similar to those of Figure \ref{fig:tetra_expt}. 
    The proposed method can reconstruct the local gradient in highly turbid media.
    }
    \label{fig:shell_expt}
\end{figure}

Figure \ref{fig:tetra_expt} and \ref{fig:shell_expt} show the results for {\it tetrapod} and {\it shell}.
In each figure, (a) shows the result obtained in clear water and (b) shows the results of the existing \cite{tsiotsios14} (second and third rows) and proposed (fourth and fifth rows).
The top row shows one of the input images.
We changed the concentration of the participating medium during these experiments (we mixed 10, 20, and 30 mL of milk with 120 L of water from left to right).
As can be seen, the result of the existing method \cite{tsiotsios14} becomes flattened as the concentration of the participating medium increases.
In contrast, the proposed method can reconstruct the detalied shape in highly turbid media.

\begin{table}[t]
  \centering
    \begin{tabular}{c|r|rrrrr} \hline
      & \cite{tsiotsios14} & 1 & 2 & 3 & 4 & 5\\ \hline 
      Error (deg.) & 19.48 & 5.96 & 4.38 & 3.62 & 3.66 & 3.66\\ \hline
    \end{tabular}
    \caption{Mean angular error of {\it sphere}. The error of the proposed method is lower than that of the backscatter-only modeling \cite{tsiotsios14}, and a few iterations are sufficient to reach convergence.}
    \label{tab:sphere_expt}
\end{table}

\section{Conclusion}
In this paper, we have proposed a photometric stereo method in participating media that considers forward scatter.
The proposed analytical model differs from the previous works in that forward scatter depends on the object's shape.
However, the shape dependency of the forward scatter makes it impossible to remove.
To address this problem, we have proposed an approximation of the large-scale dense matrix that represents the forward scatter as a sparse matrix.
Our experimental results demonstrate that the proposed method can reconstruct a shape in highly turbid media.

However, the ambiguity of the optimized support size of the kernel remains. 
We set aside an adaptive estimation of the support size for future work.

A limitation of the proposed method is that it requires a mask image of the target object.
However, in highly turbid media, it may be difficult to obtain an effective mask image.
In addition, we must initialize the object's shape, which may be solved using a depth estimation method in participating media \cite{treibitz09, asano16, dancu14}.

{\small
\section*{Acknowledgement}
This work was supported by the Japan Society for the Promotion of Science KAKENHI Grant Number 15K00237.
}

{\small
\bibliographystyle{ieee}

}

\end{document}